\documentclass[sigconf]{acmart}
\pdfoutput=1

\copyrightyear{2026}
\acmYear{2026}
\setcopyright{cc}
\setcctype{by}
\acmConference[SIGIR '26]{Proceedings of the 49th International ACM SIGIR Conference on Research and Development in Information Retrieval}{July 20--24, 2026}{Melbourne, VIC, Australia}
\acmBooktitle{Proceedings of the 49th International ACM SIGIR Conference on Research and Development in Information Retrieval (SIGIR '26), July 20--24, 2026, Melbourne, VIC, Australia}
\acmDOI{10.1145/3805712.3808402}
\acmISBN{979-8-4007-2599-9/2026/07}

\usepackage{graphicx} 
\usepackage{xcolor}
\usepackage{listings}
\usepackage{caption}
\usepackage{hyperref}
\usepackage{orcidlink}
\usepackage{enumitem}

\lstdefinestyle{custom}{
  basicstyle=\ttfamily\small,
  backgroundcolor=\color{white},
  keywordstyle=\color{blue},
  commentstyle=\color{gray},
  showstringspaces=false,
  breaklines=true,
  moredelim=**[is][\colorbox{yellow}]{@}{@}
}


\title{Unified Semantic Modeling Framework for Large-Scale Job Understanding at LinkedIn}

\author{Dan Xu}
\authornote{Equal contribution.}
\affiliation{
 \institution{LinkedIn Corporation}
 \city{Mountain View}
 \state{CA}
 \country{USA}
}
\email{dnxu@linkedin.com}

\author{Baofen Zheng}
\authornotemark[1]
\affiliation{%
  \institution{LinkedIn Corporation}
  \city{Mountain View}
  \state{CA}
  \country{USA}}
\email{bzheng@linkedin.com}

\author{Jianqiang Shen}
\affiliation{%
  \institution{LinkedIn Corporation}
  \city{Mountain View}
  \state{CA}
  \country{USA}}
\email{jshen@linkedin.com}

\author{Qi Xiao}
\affiliation{%
  \institution{LinkedIn Corporation}
  \city{Mountain View}
  \state{CA}
  \country{USA}}
\email{qixiao@linkedin.com}

\author{Benjamin Hoan Le}
\affiliation{%
  \institution{LinkedIn Corporation}
  \city{Mountain View}
  \state{CA}
  \country{USA}}
\email{ble@linkedin.com}

\author{Wen Pu}
\affiliation{%
  \institution{LinkedIn Corporation}
  \city{Mountain View}
  \state{CA}
  \country{USA}}
\email{wpu@linkedin.com}

\author{Saurabh Gupta}
\affiliation{%
  \institution{LinkedIn Corporation}
  \city{Mountain View}
  \state{CA}
  \country{USA}}
\email{saugupta@linkedin.com}

\author{Ran Zhou}
\affiliation{%
  \institution{LinkedIn Corporation}
  \city{Mountain View}
  \state{CA}
  \country{USA}}
\email{rzhou@linkedin.com}

\author{Neha Saraf}
\affiliation{%
  \institution{LinkedIn Corporation}
  \city{Mountain View}
  \state{CA}
  \country{USA}}
\email{nsaraf@linkedin.com}

\author{Alice Leung}
\affiliation{%
  \institution{LinkedIn Corporation}
  \city{Mountain View}
  \state{CA}
  \country{USA}}
\email{alleung@linkedin.com}

\author{Qianqi Shen}
\affiliation{%
  \institution{LinkedIn Corporation}
  \city{Mountain View}
  \state{CA}
  \country{USA}}
\email{qishen@linkedin.com}

\author{Liangjie Hong}
\affiliation{%
  \institution{LinkedIn Corporation}
  \city{Mountain View}
  \state{CA}
  \country{USA}}
\email{liahong@linkedin.com}

\author{Jingwei Wu}
\affiliation{%
  \institution{LinkedIn Corporation}
  \city{Mountain View}
  \state{CA}
  \country{USA}}
\email{jingwu@linkedin.com}

\author{Wenjing Zhang}
\affiliation{%
  \institution{LinkedIn Corporation}
  \city{Mountain View}
  \state{CA}
  \country{USA}}
\email{wzhang@linkedin.com}

\begin{document}

\begin{abstract}
Job understanding is critical to LinkedIn’s mission of connecting talent with opportunity. This task involves transforming unstructured and noisy job postings into standardized or derived job attributes that power numerous LinkedIn products. However, building a scalable, cost-efficient, and high-performing job understanding system remains challenging.
In this paper, we present a unified semantic modeling framework powered by a small language model (SLM) to address the challenges. 
We begin by fine-tuning an open-source SLM using a suite of carefully curated synthetic tasks augmented with reasoning traces. These tasks jointly target taxonomy-guided classification and taxonomy-agnostic entity extraction. This allows the resulting model to acquire robust zero-shot generalization for job understanding in structured and unstructured contexts. Building upon this foundation, we introduce a multi-adapter architecture with attribute grouping to facilitate efficient task-specific adaptation while streamlining model management across diverse downstream attributes.
Offline evaluations and online A/B tests demonstrate significant performance improvement while reducing operational complexity. Our work provides practical insights into building industry-scale text understanding systems.
\end{abstract}


\maketitle

\begin{table*}[tb]
\centering
\caption{Examples of important LinkedIn job attributes.}
\begin{tabular}{p{3.8cm} p{8.0cm}}
\toprule
\textbf{Category} & \textbf{Job Attributes} \\
\midrule
Large-cardinality taxonomy &
Title, Occupation, Company, Skill, Geo \\
\midrule
Small-cardinality taxonomy &
Workplace type, Employment type, Seniority, Experience level, Benefits \\
\midrule
Taxonomy-free attributes &
Salary, Years of experience \\
\midrule
Profession-specific attributes &
Nurse Specialty, Shift, Schedule, Work setting \\
\bottomrule
\end{tabular}

\vspace{2pt}
\begin{minipage}{\linewidth}
\footnotesize
\textit{Note:} Occupation is defined as a job title encompassing both a role (e.g., Engineer) and a specialty (e.g., Software). Workplace Type refers to work arrangements such as Remote, Hybrid, and Onsite. Employment Type includes categories such as Full-time, Part-time, and Contractor.
\end{minipage}

\label{tab:taxonomy}
\vspace{-6pt}
\end{table*}

\section{Introduction}
Job postings are one of LinkedIn’s most critical assets and form a foundational pillar of LinkedIn talent marketplace products.  Job understanding, which refers to the process of transforming unstructured, noisy job posting text into standardized or derived job attributes, is an essential work at LinkedIn and has been powering a wide range of applications. For example, many standardized attributes are displayed to job seekers on job detail pages to help them quickly assess the opportunities. In LinkedIn’s job search and job recommendation systems, standardized job attributes continue to play a vital role despite advancements in embedding-based methods. Attributes like company, location, occupation, seniority, and workplace type, treated as facts or important requirements, are used as essential job filters, ensuring that the job matching remains both highly precise and personalized.

Given the importance of job understanding, building and productionizing efficient and scalable machine learning (ML) models is essential. However, creating such an ecosystem  presents significant challenges. Job postings often contain long, heterogeneous text with noisy or unstructured components. This variability and complexity make accurate job understanding substantially  difficult.  Moreover, the space of job attributes is highly diverse: beyond generic attributes such as  titles and skills, LinkedIn seeks to support an expanding set of professional segments (e.g., nurses, lawyers, teachers), each characterized by its own domain-specific attributes. For instance, nurse job attributes include  specialty, shift, and license.  Table.~ref{tab:taxonomy} lists examples of important job attributes at LinkedIn based on differnt taxonomy categories. Developing models that can accurately classify such a diverse and fast-evolving attribute space at scale is non-trivial.

LinkedIn has long relied on ML models for job understanding \cite{shan_li}, typically deploying a separate model for each attribute. Those models usually required heavy feature engineering—such as text similarity signals and later embedding‑based features—and separate data pipelines to serve each model. As a result, adding a new attribute demanded substantial  effort and led to high maintenance overhead. Moreover, despite repeated iterations, traditional models often struggled to capture the semantic richness of job descriptions, resulting in suboptimal performance. Collectively, limited scalability, high maintenance, and weaker semantic modeling make traditional methods unsustainable for evolving product needs.

Large Language Models (LLMs) \cite{openai2022chatgpt, brown2020language, openai2023gpt4} have gained significant popularity for a wide range of text-based applications, including text understanding and classification \cite{sun2023carp, vajjala2025llmera, kostina2025llmclassification}. We aim at leveraging LLMs to build  an efficient semantic modeling approach for job understanding tasks, based on the premise that their strong text comprehension capabilities align well with job understanding requirements. Furthermore, LLM prompt engineering provides a practical alternative to the complex feature engineering traditionally required in earlier approaches.

However, applying LLMs to job-understanding tasks introduces additional challenges beyond those mentioned above. First, relying on commercial LLMs such as GPT-4 \cite{openai2023gpt4} is impractical due to high cost and latency. Large open‑source LLMs face similar limitations, as serving them in‑house incurs substantial latency and infrastructure costs.  To address these constraints, we target smaller size LLMs (SLMs) for job understanding tasks. Yet, SLMs often yield suboptimal performance on domain‑specific tasks; in our case, this manifests as degraded precision and recall. Furthermore, supporting a large number of heterogeneous job attributes at scale remains a significant modeling challenge.

We tackle the above challenges by developing a unified semantic modeling framework powered by SLM fine-tuning for large scale  job understanding. Specifically, we make the following contributions: 
\begin{itemize}[leftmargin=*, itemsep=0pt, topsep=2pt, label={}]
    \item We introduce an efficient methodology for fine‑tuning an open source SLM to build a job understanding base model without requiring extensive human annotation. The approach relies on carefully constructed GPT‑4–generated synthetic tasks with reasoning traces\footnote{We access GPT-4 via Azure OpenAI Services.}, enabling the model to acquire broad job semantics competence. The resulting base model shows excellent zero-shot learning performance, making fine-tuning individual tasks optional or easier.
    \item To further enhance performance for specific attributes and ensure robust attribute-level accuracy, we propose a  LoRA based \cite{hu2021lora} multi-adapter combined with a semantics‑aware attribute grouping strategy, which achieves a favorable tradeoff between performance protection and maintenance cost.
    \item We productionize the SLM base model with adapters through an efficient single‑pipeline nearline serving infrastructure, which has supported 15 key job attributes with minimal onboarding effort. Our online A/B tests demonstrated superior performance of the new job understanding model across core product metrics. 
\end{itemize}

The rest of this paper is organized as follows. Section~\ref{sec.related_work} reviews related work. Section~\ref{sec.framework} introduces the proposed semantic modeling  framework. Section~\ref{sec.experiment} presents the offline and online experimental results. Finally, Section~\ref{sec.conclusion} concludes the paper.

\section{Related Work}
\label{sec.related_work}
Machine learning approaches for job attribute standardization or extraction have been widely explored in    \cite{shan_li, careers_build_job_title, JobBERT_2021, Boselli2017, Colombo2019, Ikudo2019, Kouretsis2020, Russ2016, Zhang2022, Goindani2017, Qin2019}. The most relevant work to ours is \cite{shan_li}, which introduced LinkedIn’s job standardization models for large cardinality attributes. Their approach combines string-matching entity tagging for candidate selection with deep neural networks for ranking.   

LLMs have recently been applied to job attribute standardization \cite{LLM4Jobs, multi_label_skill_llm_2023, skill_llm_2023, occupation_skill_eval_2025}. For example, \cite{LLM4Jobs} summarizes job descriptions using an LLM, encodes the summaries and standardized occupations into embeddings, and applies embedding retrieval for occupation mapping. \cite{multi_label_skill_llm_2023} generates synthetic labeled datasets for training bi-encoder models for skill extraction, while \cite{skill_llm_2023} uses GPT to create synthetic data for binary skill classifiers, followed by GPT-based re-ranking. Finally, \cite{occupation_skill_eval_2025} proposes a framework to evaluate LLM performance on occupation classification tasks.

Prior work has focused on improving zero-shot learning in LLMs through techniques such as instruction fine-tuning \cite{wei2022finetuned} and chain-of-thought prompting \cite{zero-shot-nips2022}. In contrast, our approach targets smaller  LLMs (SLMs), enhancing their zero-shot performance via domain-specific distillation tasks.

LLM has been widely used in  query understanding \cite{pingliu_2026, luo2022query, luo2024exploring, zhang2020query, srinivasan2022quill, dai2024enhancing, abe2025llm}.  The work mainly focuses on user intention detection. In \cite{pingliu_2026}, the authors use LLMs for query attributes (i.e., search facets) tagging, but considers only a limited set of attributes, where fine-tuning a LLM with one prompt suffices to address the problem.

Our work differs from the above ones by focusing on job understanding tasks. We design a semantic modeling framework  to handle a large set of job attributes while ensuring high model performance by both improving zero-shot learning ability of the SLM base model as well as efficient task finetuning.

\section{Job Understanding Semantic Modeling Framework}
\label{sec.framework}
In this section, we introduce our proposed SLM based semantic modeling framework for job understanding. We start by providing a preliminary overview of LinkedIn job understanding tasks to establish more context for our approach.

\subsection{Preliminary on LinkedIn Job Understanding}
As mentioned earlier, LinkedIn job understanding involves transforming raw, unstructured job posting data into standardized or derived attributes. These tasks can be broadly categorized into taxonomy-based tasks and taxonomy-free tasks (e.g., salary, years of experience).
Taxonomy-based tasks can be further divided into:
\begin{itemize}[leftmargin=*, itemsep=0pt, topsep=2pt, label={}]
\item \textbf{Large-cardinality attributes} (e.g., company, location, skill, title), which traditionally \cite{shan_li} require a two-stage approach—retrieval followed by ranking—to support taxonomy scalability.
\item \textbf{Small-cardinality attributes} (e.g., workplace type, employment type, experience level), which involve direct extraction or classification without retrieval.
\end{itemize}

For small‑cardinality attributes, we include all taxonomy values in the prompt and fine‑tune an SLM to select the correct ones as a classification task. For large‑cardinality attributes, enumerating all values in the prompt is impractical, so we adopt a two‑stage design: a retrieval model first narrows the candidate set (stage‑1), and the SLM then performs classification over these retrieved candidates (stage‑2). This paper focuses on building an efficient fine‑tuned SLM for both small‑cardinality classification and stage‑2 classification for large‑cardinality attributes.

\begin{figure*}[tb]
    \centering
    \includegraphics[width=0.83\textwidth]{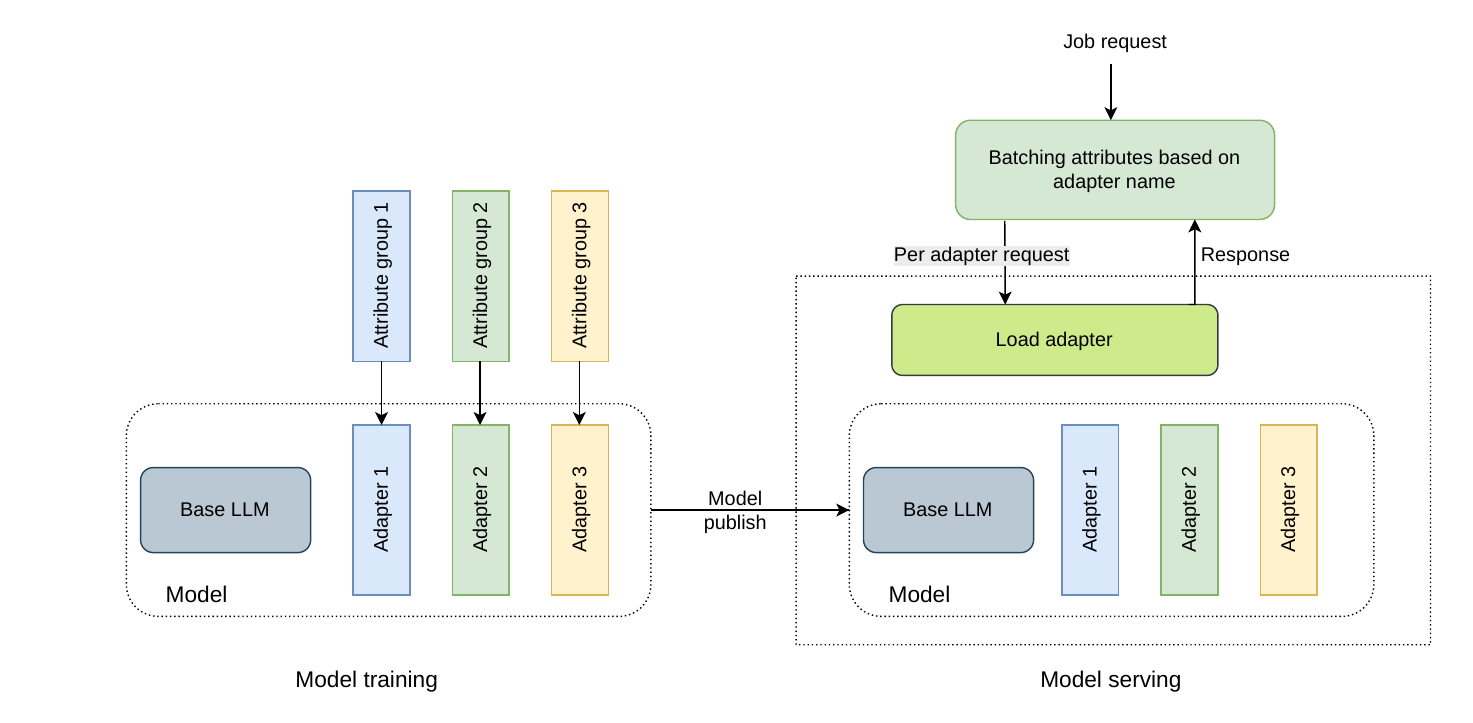}
    \vspace{-4pt}
    \caption{Multi-adapter model training and serving architecture with attribute grouping.}
    \vspace{-3pt}
    \label{fig.adapter}
\end{figure*}

\subsection{Synthetic Task Based SLM Fine-tuning}
\label{sec.framework.base}
Given the wide variety of job attribute types, our objective is to develop a base SLM with strong general zero-shot learning capabilities for job understanding tasks. This model can either directly support certain attributes or serve as a foundation for further fine-tuning on specific attributes, depending on product requirements.

We began with leveraging existing job understanding training data, primarily annotated by in-house linguistic experts, to fine-tune an open-source SLM. However, the resulting performance was unsatisfactory since high quality annotations exist only for a subset of job attributes, and the annotation data volume for each attribute is small. More importantly, training solely on annotation labels fails to capture the reasoning required for robust job comprehension, as labels convey minimal contextual information.

To accelerate training data collection without relying on annotations, we use GPT-4 to create synthetic tasks that simulate real-world job understanding scenarios and include reasoning traces to enhance domain adaptation and data efficiency. We then fine-tune SLMs using the generated task data, including instructions, labels, and explanations. Specifically, we prompt GPT‑4 to create synthetic job understanding classification task, i.e., classifying job postings into the correct taxonomy values. In this task, GPT-4 flexibly generates diverse attribute types and, for each one, constructs  a taxonomy pool-a structured set of labels enriched with definitions and aliases. These taxonomy definitions and aliases provide essential context, enabling the model to make informed classification decisions. GPT-4 then selects the appropriate labels from the created taxonomy pool and provides a reasoning paragraph explaining the choice for the given job posting. We also use GPT-4 to create a complementary taxonomy‑free entity extraction task, but due to space constraints, we omit it from this paper.

The above classification task equips the SLM with robust capabilities for taxonomy-based classification, which cover most job understanding attributes since they follow standardized taxonomies. Although the GPT-4 generated taxonomy is not fully aligned with our practical taxonomies, the inclusion of taxonomy definitions and aliases ensures high-quality synthetic taxonomies and provides essential context. The GPT‑4 generated synthetic instruction-based fine-tuning follows a similar paradigm to instruction-following model fine-tuning approaches such as Alpaca \cite{taori2023stanford}.

We sampled over 100k job postings and applied the designed prompts to generate training labels enriched with reasoning traces for fine-tuning SLMs. As demonstrated in Section \ref{sec.experiment}, this approach produced highly effective training data, enabling the fine-tuned SLMs to achieve excellent performance. The effectiveness of our large LLM–driven synthetic task approach stems from three principal factors: 1) reasoning traces and taxonomy definitions generated by the LLM enhance contextual comprehension and support more accurate inference; 2) the LLM’s ability to produce diverse attribute types ensures comprehensive coverage of job understanding tasks; 3) large-scale synthetic data, derived from extensive job posting samples, provides sufficient breadth and depth for robust knowledge acquisition and model training.

\subsection{Multi-Adapter and Attribute Grouping}
\label{sec.lora}
As Section~\ref{sec.experiment} shows, the fine-tuned base model  demonstrates strong zero-shot learning performance and can be applied directly to some job attribute understanding tasks. However, we observed that for some attributes, further fine-tuning by only a small amount of high quality linguist annotated data can yield additional improvements. This motivates us to explore task-specific fine-tuning using limited annotation data, particularly for business-critical attributes that demand  high performance.

A key challenge in task-specific fine-tuning is preserving the performance of previously optimized tasks while introducing new ones. Joint retraining across tasks is computationally expensive and leads to task interference.  To address this challenge, we adopt a multi-adapter Low-Rank Adaptation (LoRA) \cite{hu2021lora} framework in which all tasks share a frozen base model, while each task is fine-tuned independently through a lightweight adapter. Unlike other architectures such as Series and Parallel adapters \cite{wang2022adamix, he2022sparseadapter, he2022unified} that modify the Transformer structure and require full model redeployment, LoRA \cite{hu2021lora} introduces low-rank matrices on frozen weights, enabling parameter-efficient updates \cite{peft} and dynamic adapter switching. This design enables orthogonal optimization across tasks and continual extension without retraining the base model. At inference time, the system loads the base model once and dynamically activates the adapter corresponding to the request attribute, achieving efficient task specialization with minimal serving overhead.

Using LoRA adapter-based fine-tuning, we successfully improved the performance of individual attributes while preventing  regressions when adding new attribute tasks. However, as the number of attributes grows over time, e.g., those professional segment attributes, maintaining separate adapters for each attribute still introduces notable operational complexity and overhead. To address this challenge, we introduce an attribute‑grouping strategy that allows multiple semantically related attributes to share a single adapter. For each attribute, we first compute embeddings for all of its taxonomy values using an embedding SLM, and then obtain the attribute‑level representation by averaging these value‑level embeddings. We subsequently apply K‑Means clustering to the attribute embeddings to identify groups of semantically similar attributes. Attributes assigned to the same cluster share a common adapter.
In practice, attributes associated with the same professional segment (e.g., Nurse‑related attributes) are typically grouped under a shared adapter. As the taxonomy evolves and additional professional segments are  introduced, attributes from different segments may also be grouped together when their semantic representations are sufficiently similar (e.g., shift‑related attributes for Nurse and Warehouse Worker roles).
This approach significantly reduces the number of adapters while preserving task-specific performance.  Fig.~\ref{fig.adapter} illustrates the multi-adapter based training and serving architecture with attribute grouping.

\begin{figure*}[tb]
    \centering
    \includegraphics[width=0.83\textwidth]{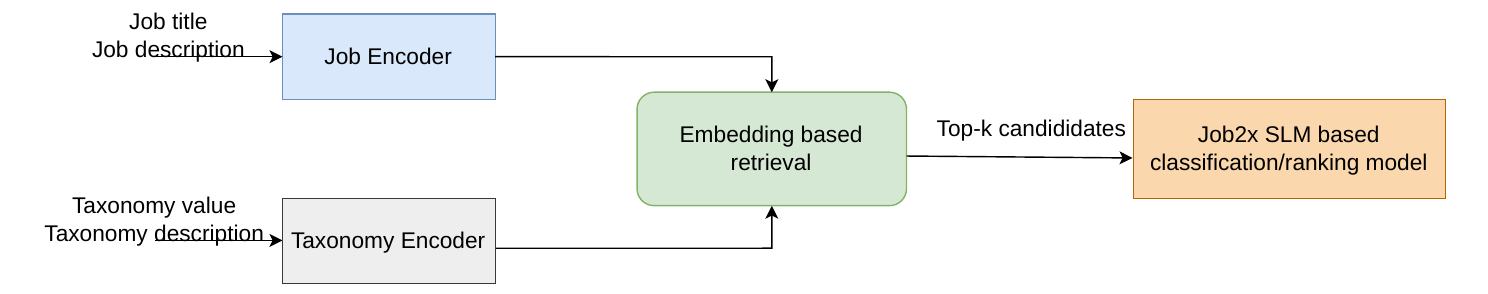}
    \caption{Job understanding two-stage model architecture for large-cardinality attributes.}
    \label{fig:two_stage}
\end{figure*}

\subsection{Support Large-Cardinality Attributes}
The above approaches have addressed most critical issues for performance assured and scalable job understanding tasks. With the fine-tuned model, we can simply write the job description and each attribute’s taxonomy values in the prompt, allowing the SLM to select the correct values. However, as mentioned, for large-cardinality attributes, a  retrieval task is needed first.

We then fine-tune an embedding-based SLM for taxonomy candidate retrieval using a Siamese bi-encoder \cite{thakur2021augmented} architecture. The retrieved  candidates are then passed to the stage-2 SLM (fine-tuned base model or adapters using the  above method) to select the final correct ones.  A key finding from our retrieval experiments is that augmenting taxonomy entries with descriptive data improves model performance, e.g., 4.7\% lift in  recall@20 for the  job occupation (Title including both specialty and role) task, underscoring the importance of rich contextual signals in candidate selection. 

The EBR model returns top-k candidates for the next stage classification model or ranking model to select or rank the final results, as illustrated in Fig.~\ref{fig.two_stage}.

Currently, among the attributes we consider, only Occupation  exhibits sufficiently large cardinality to require retrieval-based modeling; thus, scalability is not yet a limiting factor. Looking ahead, as additional large‑cardinality attributes  need to be incorporated, the same LoRA‑adapter framework with attribute grouping (Section~\ref{sec.lora}) can be extended to the embedding‑retrieval model to maintain scalability.

\section{Experimental Results}
\label{sec.experiment}



\subsection{Offline evaluation}

\begin{table}[t]
\centering
\caption{Precision (P) and Recall (R) of  fine-tuning strategies}
\setlength{\tabcolsep}{6pt}
\begin{tabular}{lcccc}
\toprule
\textbf{Task / P/R} & \textbf{No FT} & \textbf{COMB} & \textbf{SYNTH} & \textbf{Adapter} \\
\midrule
Specialty & 62\% / 45\% & 82\% / 24\% & 89\% / 64\% & 91\% / 82\% \\
Shift     & 35\% / 27\% & 28\% / 21\% & 93\% / 85\% & 96\% / 94\% \\
\bottomrule
\end{tabular}

\vspace{-6pt}
\small
\label{tab:offline_metric1}
\end{table}

We use Flan‑T5‑XL \cite{flan-t5} as our backbone SLM and evaluate the proposed modeling framework on human‑annotated datasets. We observe comparable performance across other SLMs when applying the same  methodology.

\begin{itemize}[leftmargin=*, itemsep=0pt, topsep=2pt, label={}]
\item \textbf{No FT}: No fine-tuning applied to the Flan‑T5‑XL model.
\item \textbf{COMB}: Joint fine-tuning on multiple job attribute tasks using existing annotated data (e.g., title, skill, seniority, salary, etc).
\item \textbf{SYNTH}: Fine-tuning the Flan‑T5‑XL SLM by the GPT-4-generated synthetic task data, including labels and reasoning traces (Section~\ref{sec.framework.base}). Compared to COMB, SYNTH provides greater attribute diversity and a substantially larger volume of training data.
\item \textbf{Adapter}: Further fine-tuning on top of the SYNTH approach created model using a LoRA adapter \cite{hu2021lora} and a small set of linguist-annotated data for each specific task.
\end{itemize}

In Table~\ref{tab:offline_metric1}, we evaluate four strategies on two critical Nurse attributes—\textit{Specialty} and \textit{Shift}. The Nurse segment represents one of LinkedIn’s highest-priority professional groups for business growth.
 To assess generalization, COMB and SYNTH were trained without any Nurse related tasks. Results show that COMB fails to improve generalization, while SYNTH achieves significant gains. The Adapter approach delivers the best performance, demonstrating the effectiveness of task-specific fine-tuning with LoRA adapters built on the SYNTH-based model. Notably, the amount of annotation data required for  the task-specific fine-tuning is small. For the two Nurse attributes,  we find that approximately 300 samples are sufficient to achieve the performance shown in Table~\ref{tab:offline_metric1}. These results and findings highlight the strong zero-shot learning capabilities of our fine-tuned base model, substantially reducing the need for extensive task-specific training or large volumes of labeled data.

\begin{table}[tb]
\centering
\caption{Comparison of precision and recall (P/R) between legacy models and the fine-tuned SLM. }
\begin{tabular}{lcc}
\toprule
\textbf{Attribute} & \textbf{Legacy Model} & \textbf{Fine-tuned SLM} \\
\midrule
Occupation      & 71\% / 25\% & 83\% / 83\% \\
Seniority       & 70\% / 70\% & 84\% / 84\% \\
Workplace Type  & 95\% / 29\% & 98\% / 97\% \\
\bottomrule
\end{tabular}
\label{tab:performance_comparison}
\vspace{-2pt}
\end{table}

In Table~\ref{tab:performance_comparison}, we compare the developed SLM model performance to legacy models for some other business critical attributes, including Occupation, Seniority, and Workplace Type.   For Occupation, the baseline is a GNN-job-encoding \cite{li2024jobembedding, liu2025linksage} based deep model; for Seniority, a deep MLP model; and for Workplace Type, a deep model leveraging job representations encoded using BERT \cite{devlin2019bert}. Note that the Occupation evaluation data  was collected only from the challenging cases where the job title  model failed to produce a valid occupation.

\begin{table}[tb]
\centering
\caption{Product metrics improvements via online A/B tests.}
\begin{tabular}{l p{5cm}}
\toprule
\textbf{Attributes} & \textbf{Talent Marketplace Top-tier metrics} \\
 \hline
Occupation    & +0.05\% WAU, +0.15\% Job Sessions, +0.49\% QA, -9.25\% JRF.    \\
Seniority    & +0.38\% QA, -2.9\% JD2A  \\
Workplace type    &  +0.48\% QA  \\
Nurse attributes    & +0.62\% Job WAU for Nurse segment  \\
\bottomrule
\end{tabular}
\label{tab:online_metric}
\vspace{-6pt}
\end{table}

\subsection{Model Serving}

Job understanding tasks are primarily served in a nearline fashion, processing streaming LinkedIn job posting data.  We built an efficient serving infrastructure that integrates a unified streaming processor, a gRPC based mid‑tier service, and GPU‑backed model hosts. For attributes served by the base model or those sharing the same adapter, a single model invocation retrieves all corresponding attribute values. Attributes requiring different adapters involve an adapter switch; however, the associated time overhead is negligible. This unified approach significantly improves service efficiency and reduces maintenance costs by eliminating previous redundant pipelines and simplifying operational complexity.

We have deployed over 50 NVIDIA A100 GPUs to serve LinkedIn job traffic in production. As many jobs share identical  job descriptions—for example, roles from the same employer or jobs that are re‑ingested—we implemented a caching layer that significantly reduces the volume of requests invoking the job understanding SLM. Under the current configuration, a single GPU sustains approximately 1 job QPS, with a median (P50) latency of about 0.8 seconds and a P95 latency around 1.2 seconds when serving the Flan‑T5‑XL model. Note that the input prompt length averages approximately 400 tokens, depending on job description length, while the generated output typically contains no more than 5 tokens. Compared to commercial LLMs (e.g., GPT‑4/5), which incur token‑based usage costs, provisioning a fixed number of GPUs results in substantially lower and more predictable operational expenses.  

Furthermore, we are planning to migrate from the encoder\textendash\allowbreak
decoder\textendash\allowbreak based Flan‑T5‑XL model to more efficient small
decoder\textendash\allowbreak only models, such as Qwen3‑1.7B~\cite{qwen3technicalreport}, which can be
served via vLLM. This architectural transition is expected to further reduce GPU
provisioning costs and end‑to‑end inference latency through advanced serving
optimizations, including continuous batching and prefix caching.

\subsection{Online A/B testing results}
Based on the same serving infrastructure, we conducted online A/B tests comparing the newly developed SLM with the legacy models across key LinkedIn products, including job search and job recommendations. The A/B tests demonstrated statistically significant improvements in top‑tier product metrics ($p$<.05) for the new models, including user‑engagement metrics, e.g., Weekly Active Users (WAU), Job WAU, Job Sessions, Qualified Applications (QA), and relevance quality related metrics, e.g., Job Recommendation Facepalms (JRF) and Job Dismiss‑to‑Application Ratio (JD2A), as summarized in Table~\ref{tab:online_metric}.  The improvements stem from the higher‑precision and higher‑recall job attributes produced by the new job understanding models, which enable the delivery of more relevant, higher‑quality, and more accurately matched jobs to job seekers.

\section{Conclusion}
\label{sec.conclusion}
In this paper, we propose a unified semantic modeling framework based on efficient SLM fine-tuning for large-scale job understanding. Our approach introduces an efficient methodology to build a base model with strong zero-shot learning capabilities, leveraging GPT-4 generated synthetic tasks. Furthermore, we use a multi-adapter framework to support flexible task‑specific fine‑tuning while keeping the number of adapters manageable through semantics‑aware attribute grouping.   Experimental results show that our method significantly outperforms legacy models in both offline evaluation  and online A/B testing. Our contributions not only address fundamental limitations of traditional job understanding modeling and serving stacks, but also provide practical insights into building industrial-scale text understanding systems.



\bibliographystyle{ACM-Reference-Format}
\bibliography{main}

\end{document}